\pdfoutput=1
\documentclass[10pt]{article}

\usepackage[margin=1in]{geometry}
\usepackage[T1]{fontenc}
\usepackage{newtxtext,newtxmath}
\usepackage{microtype}
\usepackage{amsmath}
\usepackage{booktabs}
\usepackage{tabularx}
\usepackage[retain-explicit-plus]{siunitx}
\usepackage{graphicx}
\usepackage[font=small,labelfont=bf,skip=4pt]{caption}
\usepackage{subcaption}
\usepackage[round]{natbib}
\usepackage{titlesec}
\usepackage{xcolor}
\usepackage{tikz}
\usetikzlibrary{arrows.meta,positioning}
\usepackage{enumitem}
\usepackage{xurl}
\usepackage[hidelinks]{hyperref}
\usepackage[capitalise,noabbrev]{cleveref}

\setlist{nosep,leftmargin=*}
\titlespacing*{\paragraph}{0pt}{0.8ex plus 0.4ex minus 0.2ex}{0.8em}
\hypersetup{pdftitle={Good Judgment, Bad Advice: Decision-Model Controllers Raise Agent Cost Without Improving Success}, pdfauthor={Naman Vats}}

\title{Good Judgment, Bad Advice: Decision-Model Controllers\\Raise Agent Cost Without Improving Success}
\author{Naman Vats\\\texttt{be10360.17@bitmesra.ac.in}}
\date{}

\begin{document}
\maketitle

\begin{abstract}
Agent scaffolds increasingly delegate small decisions (whether to stop, verify, or change approach) to cheaper models.
We ask whether a typed decision model, Jev~1.13, consulted only at deterministic decision points, makes a Claude Sonnet~5.5 terminal agent cheaper without making it worse.
In a pre-specified study on Terminal-Bench~2.1, we hold one agent loop fixed and vary only the controller: none, deterministic rules, a small LLM (Claude Haiku~5.5) answering the same questions, Jev, and a placebo that samples Jev's development-phase choice frequencies at the same moments.
On 21 held-out tasks (315 trials), Jev raised cost per attempt by 19\% (geometric-mean ratio 1.19, 95\% CI 1.02--1.36) at an identical pass rate (73.0\% vs.\ 73.0\%).
Jev was cheaper than the small-LLM controller (ratio 0.81) but 22\% more expensive than the placebo, and no pass-rate difference between arms was detectable.
Offline, Jev ranked finish requests better than the alternatives (AUROC 0.78 vs.\ 0.62 for rules and 0.60 for the small LLM) but was poorly calibrated: its mean probability that a run was complete was 0.50 when 87\% of those runs passed, so it asked the agent to verify work that was already correct.
No controller ever chose to abandon a run.
Informative decision signals did not become useful interventions; we discuss what that would require.
Code: \url{https://github.com/namanvats/decision-model-controllers}.
\end{abstract}

\section{Introduction}

Terminal agents solve tasks by running long loops of model calls and shell commands~\citep{yao2023react,yang2024sweagent}.
Their cost is dominated by the reasoning model, and scaffolds add control logic around it: when to stop, when to check work, when to change approach.
A natural way to make that logic both smarter and cheaper is to delegate it to a small model.
Typed decision models are built for this: given a state and a fixed set of options, they return a choice and a probability distribution, at a small fraction of the price of a reasoning model.
Jev~1.13~\citep{jevdocs}, for example, is priced at \$0.042 per million input tokens; in our runs a decision cost about \$0.00008.

A controller can only save money if its judgments carry information about how the run will end \emph{and} its actions turn that information into savings.
Evaluations of judgment models mostly measure the first~\citep{zheng2023judging,kadavath2022know}.
We test both, end to end, with the baselines that separate them: deterministic rules (is a model needed at all?), a small LLM asked the same questions (why not a cheap LLM?), and a placebo that intervenes at the same moments with random choices drawn from Jev's own choice frequencies (does the decision content matter, or only the nudge?).
Our contributions are a pre-specified, placebo-controlled design that varies only the controller of a fixed agent loop (\cref{sec:design}); a confirmatory result that Jev raised cost by 19\% without changing the pass rate (\cref{sec:c1}); comparisons with the small LLM, rules and placebo (\cref{sec:secondary}); and an account of why, from offline judgment quality and the logged decisions (\cref{sec:offline,sec:mechanism}).

\section{Related work}

ReAct-style loops interleave reasoning with tool calls~\citep{yao2023react}, and the interface around the model matters as much as the model~\citep{yang2024sweagent}.
Reflexion adds verbal self-feedback between attempts~\citep{shinn2023reflexion}; by contrast, asking a model to re-check its work without new external information often fails to improve it~\citep{huang2024selfcorrect}; our verification nudge is of this kind, although the agent can gather new evidence by running commands.
Cascades and routers send queries to cheaper models when possible~\citep{chen2023frugalgpt,ong2024routellm}; we study a different lever, where the reasoning model is fixed and a cheap model decides only how the loop proceeds.
Selective prediction and calibration study when confidence can be acted on~\citep{geifman2017selective,guo2017calibration}, LLMs can estimate whether their own answers are correct~\citep{kadavath2022know}, and LLM judges are widely used to grade outputs~\citep{zheng2023judging}.
We measure ranking and calibration offline, and then whether acting on them helps online.

\section{Study design}
\label{sec:design}

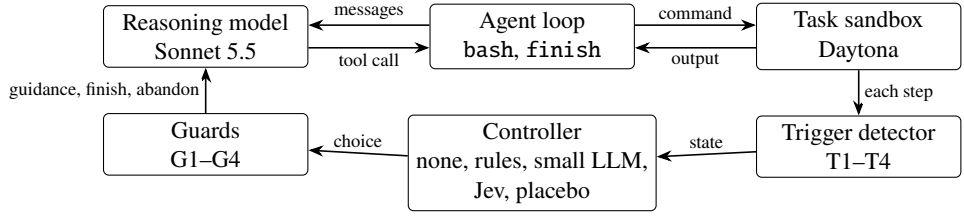
\begin{figure}[t]
  \centering
  \begin{tikzpicture}[
      node distance=6mm and 16mm,
      box/.style={draw, rounded corners=2pt, align=center, font=\small, inner sep=3pt,
                  minimum height=8mm, minimum width=27mm},
      arr/.style={-{Stealth[length=2mm]}, semithick},
      lab/.style={font=\scriptsize, fill=white, inner sep=1pt}]
    \node[box] (model) {Reasoning model\\Sonnet~5.5};
    \node[box, right=of model] (loop) {Agent loop\\\texttt{bash}, \texttt{finish}};
    \node[box, right=of loop] (sandbox) {Task sandbox\\Daytona};
    \node[box, below=of sandbox] (triggers) {Trigger detector\\T1--T4};
    \node[box, below=of loop] (controller) {Controller\\none, rules, small LLM,\\Jev, placebo};
    \node[box, below=of model] (guards) {Guards\\G1--G4};
    \draw[arr] ([yshift=1.5mm]loop.west) -- node[lab, above=1pt] {messages} ([yshift=1.5mm]model.east);
    \draw[arr] ([yshift=-1.5mm]model.east) -- node[lab, below=1pt] {tool call} ([yshift=-1.5mm]loop.west);
    \draw[arr] ([yshift=1.5mm]loop.east) -- node[lab, above=1pt] {command} ([yshift=1.5mm]sandbox.west);
    \draw[arr] ([yshift=-1.5mm]sandbox.west) -- node[lab, below=1pt] {output} ([yshift=-1.5mm]loop.east);
    \draw[arr] (sandbox.south) -- node[lab, right=1pt] {each step} (triggers.north);
    \draw[arr] (triggers) -- node[lab, above=1pt] {state} (controller);
    \draw[arr] (controller) -- node[lab, above=1pt] {choice} (guards);
    \draw[arr] (guards.north) -- node[lab, left=1pt] {guidance, finish, abandon} (model.south);
  \end{tikzpicture}
  \caption{The fixed agent loop. Triggers are detected and logged in every arm; only the controller differs. Its choice passes through guards before it can append a fixed guidance message to the conversation or end the run.}
  \label{fig:loop}
\end{figure}

\paragraph{Agent loop.}
One loop is shared by all arms (\cref{fig:loop}).
Each step is one call to the reasoning model, which may invoke \texttt{bash} or \texttt{finish}; a reply without a tool call counts as a finish request.
Commands run through Harbor~\citep{harbor} in a fresh shell inside a Daytona sandbox~\citep{daytona} built from the task's own image.
Tool output is truncated to its first 4{,}000 and last 8{,}000 characters, the context is never compacted, and the loop and all API keys stay on the host.

\paragraph{Decision points.}
Four deterministic triggers are detected in every arm, including the one without a controller, and the state at each is logged:
T1, repeated failure (two or more consecutive failing commands);
T2, no progress (a repeated command or repeated output);
T3, a finish request;
T4, a checkpoint every 10 steps.
At most one trigger fires per step, and a trigger type cannot refire within three steps (exact definitions in \cref{app:params}).

\paragraph{Questions and actions.}
At T1, T2 and T4 the controller answers \emph{Q-progress} with \texttt{continue}, \texttt{investigate}, \texttt{recover} or \texttt{abandon}; at T3 it answers \emph{Q-finish} with \texttt{finish} or \texttt{verify}.
\texttt{investigate}, \texttt{recover} and \texttt{verify} append one of three fixed advisory messages (\cref{app:prompts}) that never contain commands, paths or test information; \texttt{abandon} ends the run as a failure and \texttt{finish} ends it normally.
The controller sees only a compact state built from the agent's transcript: the instruction, the trigger, step and spend against their caps, the last five commands with exit codes and output tails, failure and repetition counters, whether a verification-like command ran after the last edit, the finish summary, and its own earlier choices.
It never sees tests, verifier output or reference solutions.

\paragraph{Controllers.}
\emph{Native} detects and logs triggers but never consults a controller.
\emph{Rules} answers \texttt{recover} at T1 and T2, \texttt{continue} at T4, \texttt{verify} at T3 unless a verification-like command ran after the last edit, and \texttt{abandon} after four consecutive failures once 40\% of the step budget is used.
The \emph{small LLM} is Claude Haiku~5.5 with reasoning disabled; it receives the same state and option descriptions as Jev and must return a single choice under a JSON schema.
\emph{Jev} is \texttt{typesafe/jev-1.13} (served snapshot \texttt{jev-1.13-20260917}) asked a \emph{choice} question with the same descriptions.
The \emph{placebo} samples from Jev's development-phase frequencies (Q-progress: continue 0.625, investigate 0.125, recover 0.25, abandon 0; Q-finish: finish 0.652, verify 0.348), seeded per trial.

\paragraph{Guards.}
Guards override every controller.
G1 enforces hard limits on steps, spend, tokens and wall time.
G2 allows \texttt{abandon} only once 40\% of the step budget is used and the run shows a failure streak or repeated lack of progress; otherwise it becomes \texttt{recover}.
G3 accepts the third finish request after two \texttt{verify} answers.
G4 replaces a timeout, transport error or malformed answer with \texttt{continue} or \texttt{finish}, and logs it.

\paragraph{Models and infrastructure.}
All models are called through OpenRouter~\citep{openrouter}.
Sonnet~5.5 runs with reasoning effort \emph{high} (reasoning is mandatory for this model) and prompt caching; no Sonnet~5.5 endpoint accepts a temperature parameter, so sampling is not greedy.
Hosts are chosen by OpenRouter among endpoints that support every request parameter; of 4{,}780 main-model calls in the evaluation, 94.6\% were served by Claude Platform on AWS and 5.4\% by Azure, with a similar mix in every arm.
Cost is the provider-reported cost of every reasoning-model and controller call; sandbox compute is excluded.
Harbor~0.24.0 runs the task's verifier after the agent stops and enforces each task's own agent timeout (at most 60 minutes for the tasks used); a reward of 1 is a pass.

\paragraph{Protocol.}
Terminal-Bench~2.1~\citep{tbench21} has 89 tasks; we pinned the hub dataset by content digest and excluded two tasks whose agent timeouts exceed 60 minutes and one that Sonnet~5.5 refuses on its first turn (found during calibration, before the split).
\emph{Calibration:} 30 tasks sampled across categories were run twice with native under generous caps (100 steps, \$3), stratified as \emph{hard} (0/2 passes), \emph{mid} (1/2) or \emph{easy} (2/2), and split into 8 development tasks (6 easy, 1 mid, 1 hard) and 21 held-out tasks (14 easy, 3 mid, 4 hard); calibration attempts are not used in any result.
\emph{Caps} were set to 60 steps and \$1.00 per trial for every arm, above every calibration run (passing runs: at most 52 steps and \$0.48; all runs: at most \$0.98).
\emph{Development:} the four non-placebo arms ran twice on the development tasks; nothing was tuned, and the runs only supplied the placebo distribution.
\emph{Evaluation:} with the held-out list checksummed and every configuration fingerprinted, 21 tasks $\times$ 5 arms $\times$ 3 attempts gave 315 trials, with arm order randomized within each task--attempt block.
No trial failed for infrastructure reasons; model spend was \$59.46 for the evaluation and \$72.29 in total.

\paragraph{Statistics.}
For each task and arm we average over attempts and compare arms paired by task, resampling tasks 10{,}000 times~\citep{efron1993bootstrap} for percentile 95\% intervals.
The cost effect is the geometric mean over tasks of the per-task ratio of USD per attempt.
The confirmatory comparison C1 (Jev vs.\ native) tests the cost ratio and checks the pass-rate difference against a pre-specified margin of $-10$ percentage points (pp) using the one-sided 95\% lower bound.
C2--C4 compare Jev with the small LLM, rules and placebo, with Holm-adjusted bootstrap $p$-values~\citep{holm1979simple}.
\Cref{sec:offline,sec:mechanism} are exploratory.

\section{Results}

\begin{table}[t]
  \centering
  \caption{Per-arm results on the 21 held-out tasks (63 attempts per arm). Decisions: controller consultations per attempt; verify: verification messages appended per attempt.}
  \label{tab:summary}
  \begin{tabular}{lrrrrrrr}
\toprule
Arm & Pass (\%) & USD/attempt & USD/solved & Steps & Wall (min) & Decisions & Verify \\
\midrule
native & 73.0 & 0.177 & 0.243 & 13.3 & 9.4 & -- & -- \\
rules & 73.0 & 0.191 & 0.262 & 15.6 & 9.5 & 3.14 & 0.92 \\
small LLM & 76.2 & 0.213 & 0.280 & 17.8 & 10.9 & 3.65 & 1.30 \\
Jev & 73.0 & 0.206 & 0.283 & 16.0 & 10.1 & 2.79 & 0.51 \\
placebo & 71.4 & 0.156 & 0.218 & 13.2 & 9.3 & 2.32 & 0.33 \\
\bottomrule
\end{tabular}

\end{table}

\begin{table}[t]
  \centering
  \caption{Paired comparisons (by task, 21 tasks). Pass-rate differences in percentage points; cost as geometric-mean ratio of USD per attempt. $p$: bootstrap $p$-value for the cost ratio, unadjusted for C1 (confirmatory) and Holm-adjusted across C2--C4. All Holm-adjusted pass-rate $p$-values are 1.00.}
  \label{tab:comparisons}
  \begin{tabular}{llrcrcr}
\toprule
& Comparison & $\Delta$ pass & 95\% CI & Cost ratio & 95\% CI & $p$ \\
\midrule
C1 & Jev vs.\ native & \num{+0.0} & [\num{-11.1}, \num{+12.7}] & 1.19 & [1.02, 1.36] & 0.026 \\
C2 & Jev vs.\ small LLM & \num{-3.2} & [\num{-11.1}, \num{+4.8}] & 0.81 & [0.68, 0.95] & 0.019 \\
C3 & Jev vs.\ rules & \num{+0.0} & [\num{-7.9}, \num{+9.5}] & 0.96 & [0.83, 1.11] & 0.623 \\
C4 & Jev vs.\ placebo & \num{+1.6} & [\num{-4.8}, \num{+9.5}] & 1.22 & [1.08, 1.39] & 0.001 \\
\bottomrule
\end{tabular}

\end{table}

\subsection{Confirmatory: Jev versus no controller}
\label{sec:c1}

Jev increased cost per attempt by 19\% (geometric-mean ratio 1.19, 95\% CI 1.02--1.36; \cref{tab:comparisons}).
The pass rate was identical, 73.0\% in both arms (difference $+0.0$\,pp, 95\% CI $-11.1$ to $+12.7$).
The one-sided 95\% lower bound, $-9.5$\,pp, lies just inside the $-10$\,pp margin; with 21 tasks this is at the limit of what the design can resolve, so we read it as no evidence of a large drop rather than as noninferiority.
The hypothesis that Jev lowers cost is rejected.

\subsection{Secondary: small LLM, rules and placebo}
\label{sec:secondary}

\begin{figure}[t]
  \centering
  \begin{subfigure}[t]{0.49\linewidth}
    \centering
    \includegraphics[width=\linewidth]{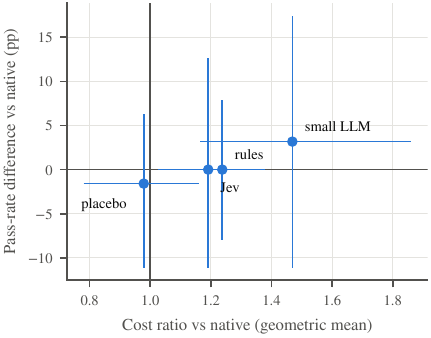}
    \caption{Each controller against native: paired cost ratio and pass-rate difference with 95\% bootstrap intervals.}
    \label{fig:effects}
  \end{subfigure}\hfill
  \begin{subfigure}[t]{0.49\linewidth}
    \centering
    \includegraphics[width=\linewidth]{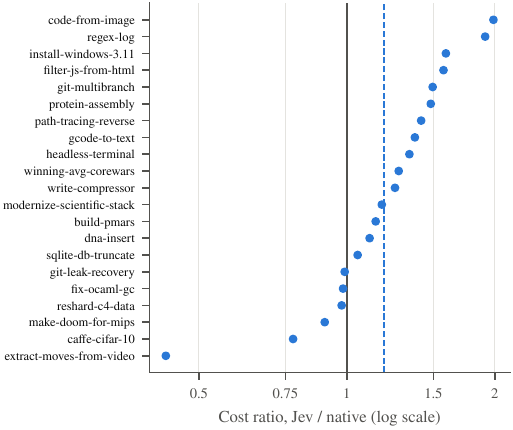}
    \caption{Per-task cost ratio of Jev to native; dashed line: geometric mean (1.19).}
    \label{fig:forest}
  \end{subfigure}
  \caption{Informed controllers raise cost; the placebo does not. The increase is spread across most tasks.}
  \label{fig:results}
\end{figure}

Jev cost 19\% less than the small LLM making the same decisions and 22\% more than the placebo, and matched the rules arm; no pass-rate difference was detectable (\cref{tab:comparisons}).
Against native (\cref{fig:effects}), all three informed controllers raised cost (rules $\times$1.24, 95\% CI 1.12--1.38; small LLM $\times$1.47, 1.16--1.86; Jev $\times$1.19), whereas the placebo did not ($\times$0.98, 0.78--1.16).

\subsection{Offline decision quality}
\label{sec:offline}

To separate judgment from action, we replayed every trigger state logged by native in the evaluation to each controller, without acting on the answers, and scored them against the trial's grader outcome (\cref{tab:offline}).
For Q-finish the score is the probability of \texttt{finish} and the label is a pass; for Q-progress it is the probability of \texttt{abandon} and the label is a failure; rules and the small LLM return a single choice, scored as 1 or 0.

Jev ranked finish requests best (AUROC 0.78, against 0.62 for rules and 0.60 for the small LLM), but it was miscalibrated toward ``not done'': its mean probability of \texttt{finish} was 0.50, while 87\% of these runs passed.
It therefore chose \texttt{verify} for 29 of 53 finish requests, and its argmax accuracy (0.55) fell below the base rate (0.87).
The ranking is real: runs given $P(\texttt{finish})\geq0.5$ passed 96\% of the time (23 of 24) and runs below 0.5 passed 79\% (23 of 29), so even the runs Jev doubted most were mostly correct.
On Q-progress no controller ever chose \texttt{abandon}, and Jev's probability of abandoning barely separated failing from passing runs (AUROC 0.55).
The 111 states logged during calibration showed the same pattern (finish AUROC 0.77 for Jev and 0.46 for the small LLM), although they predate the final verification feature.

\begin{table}[t]
  \centering
  \caption{Offline decision quality on states logged by native in the evaluation, labelled with the trial outcome. Accuracy is that of the argmax choice; no controller answer fell back.}
  \label{tab:offline}
  \begin{tabular}{lrrr}
\toprule
Controller & AUROC & ECE & Accuracy \\
\midrule
\multicolumn{4}{l}{\emph{Q-finish} ($n=53$ states)} \\
\quad base rate & 0.50 & 0.00 & 0.87 \\
\quad rules & 0.62 & 0.55 & 0.45 \\
\quad small LLM & 0.60 & 0.70 & 0.30 \\
\quad Jev & 0.78 & 0.37 & 0.55 \\
\multicolumn{4}{l}{\emph{Q-progress} ($n=72$ states)} \\
\quad base rate & 0.50 & 0.00 & 0.64 \\
\quad rules & 0.50 & 0.64 & 0.36 \\
\quad small LLM & 0.50 & 0.64 & 0.36 \\
\quad Jev & 0.55 & 0.63 & 0.36 \\
\bottomrule
\end{tabular}

\end{table}

\subsection{Where the extra cost comes from}
\label{sec:mechanism}

\paragraph{Verification and longer runs.}
Cost rose with how often a controller asked the agent to verify (\cref{tab:summary}): placebo 0.33 verification messages per attempt (cost ratio 0.98), Jev 0.51 (1.19), rules 0.92 (1.24) and small LLM 1.30 (1.47).
Verification rounds alone do not account for Jev's increase: of its extra \$0.029 per attempt, \$0.011 was spent after the first verification message.
The rest came from longer trajectories throughout (16.0 steps per attempt against 13.3), consistent with its progress messages (Jev chose \texttt{investigate} or \texttt{recover} at 30 of 92 progress decisions), although this part of the decomposition is correlational.

\paragraph{Spread across tasks.}
Jev was more expensive on 15 of 21 tasks (\cref{fig:forest}), and removing any single task leaves the ratio between 1.16 and 1.25.
Of the six cheaper tasks, three were within 3\% of native; the three clear reductions were hard tasks whose runs mostly ended at Harbor's timeout or the budget cap, where there is usually no finish request and therefore nothing to verify.
Pass rates differed between Jev and native on only three tasks (\cref{app:pertask}); easy tasks passed almost always and hard tasks almost never.

\paragraph{Abandon was never used.}
No controller chose \texttt{abandon} in any of the 315 trials, so G2 never had to block one; the placebo could not abandon by construction.
The only action designed to save money played no part, and hopeless runs ran until a limit: all 15 runs of \texttt{make-doom-for-mips} failed at Harbor's timeout (11) or the \$1.00 cap (4).

\paragraph{Overhead.}
Controller calls were negligible: Jev averaged 0.52\,s per decision and 0.11\% of spend, the small LLM 1.46\,s and 0.47\%; no call timed out or returned a malformed answer.

\section{Discussion}

\paragraph{Why informed control did not pay.}
Three properties of this setting work against an advisory controller.
First, the agent's own finish requests were usually right (87\% of native finish requests passed), so a verification round is mostly spent re-checking correct work.
Second, Jev's probabilities rank well but sit far below the base rate, so acting on its argmax converts a good ranking into frequent, low-value interventions; a base-rate-aware threshold would intervene far less, but even Jev's least confident group passed 79\% of the time.
Third, most failures (10 of 17 for native) ended at a timeout or a cap rather than with a finish request, and no controller proposed to abandon them; the abandon guard was never the obstacle.

\paragraph{What the placebo shows.}
Random guidance at the same moments matched every informed controller's pass rate without their cost.
Without the placebo, the comparison against native alone would suggest only that Jev is somewhat expensive; with it, the evidence says that Jev's choices, not the act of intervening, add the cost.

\paragraph{Implications.}
Controllers for agent loops should be evaluated end to end on cost, against rules and a placebo, not only on judgment accuracy.
Typed decision models are cheap and fast enough that overhead is not the problem (Jev cost 19\% less than the small-LLM controller and answered about three times faster); what matters is the policy that maps their probabilities to actions.
At scale, a 19\% increase in cost per attempt applies to every task an agent runs.
A follow-up should act on calibrated probabilities with base-rate-aware thresholds and give the controller stopping actions it will actually use, pre-specified on new tasks.
Two lessons for evaluation harnesses also emerged: a naive ``did the agent verify?'' feature misses agents that check work by running their own programs, and safety refusals return an empty reply that a loop can mistake for a finish request.

\section{Limitations}

\begin{itemize}
  \item \emph{Sample.} The held-out set has 21 tasks, 14 of them easy and only 3 mid, so pass-rate intervals are wide (about $\pm$11--13\,pp) and the C1 guardrail is near the design's resolution. The cost result is robust (15 of 21 tasks; leave-one-out 1.16--1.25), but the pass-rate result is weak evidence either way.
  \item \emph{One agent and one decision model.} We study one reasoning model in one minimal loop, with Jev~1.13 and our option wording and messages. A pre-specified replication on an open-weight model was not run: its OpenRouter hosts serve different quantizations, so the model could not be held fixed across arms.
  \item \emph{Nondeterminism and hosts.} Sonnet~5.5 accepts no temperature parameter and hosts were not pinned; attempts are replicates, and the host mix was similar across arms.
  \item \emph{Placebo.} Its distribution comes from a small development sample (8 progress and 23 finish decisions) and gives abandon zero probability; it tests decision content, not intervention frequency.
  \item \emph{Strata and labels.} Strata come from two calibration attempts, and offline labels use the trial outcome rather than the counterfactual outcome of a different decision.
\end{itemize}

\section{Conclusion}

A cheap, fast typed decision model consulted at well-defined decision points did not make a strong terminal agent cheaper: it raised cost by 19\% at an unchanged pass rate, and a placebo matched its pass rate for less.
Its judgments were informative but miscalibrated, its interventions mostly re-checked correct work, and the one cost-saving action it could take was never chosen.
The gap between an informative signal and a useful intervention is the policy in between, and that policy should be tested with placebo controls and cost as the primary outcome.

\paragraph{Reproducibility.}
The benchmark version, Harbor version, model identifiers, caps, trigger and guard parameters, and all prompts are given in the paper and its appendices; every trial records its configuration fingerprint, lockfile hash and Harbor command.
The harness, configurations and analysis scripts are publicly available.\footnote{\url{https://github.com/namanvats/decision-model-controllers}}

\bibliographystyle{plainnat}
\bibliography{refs}

\clearpage
\appendix

\section{Prompts and messages}
\label{app:prompts}

\paragraph{System prompt (reasoning model).}
\begin{quote}\small
You are an autonomous agent working in a Linux terminal to complete a task.\par
You have two tools:\par
\texttt{bash}: run a shell command and see its exit code and output. Each call runs in a fresh shell: the working directory and environment variables do not persist between calls, so use absolute paths or chain commands with \texttt{\&\&}. Commands run non-interactively; avoid commands that wait for input. Long-running commands are killed after \texttt{timeout\_s} seconds (exit code 124).\par
\texttt{finish}: call this when the task is complete, with a short summary of what you did and how you checked it.\par
Work step by step. Inspect the environment before changing it, and verify your work before calling \texttt{finish}.
\end{quote}

\paragraph{Guidance messages.}
Each is prefixed with ``[controller guidance --- advisory]''.
\begin{description}
  \item[I (investigate)] You may be acting without enough information about the environment. Before making further changes, gather the facts you need: inspect the relevant files, tools, versions, and the exact requirements in the task.
  \item[R (recover)] Your current approach appears to be failing repeatedly. Step back, identify why it is failing, and try a different approach rather than repeating the same commands.
  \item[V (verify)] The requirements may not be fully checked yet. Before finishing, verify each explicit requirement in the task against the actual state of the environment, then finish only if they are met.
\end{description}

\paragraph{Option descriptions (shared by Jev and the small LLM).}
\begin{center}\small
\begin{tabularx}{\linewidth}{@{}llX@{}}
  \toprule
  Question & Option & Description \\
  \midrule
  Q-progress & \texttt{continue} & The agent is making progress toward the task goal. \\
             & \texttt{investigate} & The agent is acting without enough information about the environment. \\
             & \texttt{recover} & The agent is stuck repeating a failing approach. \\
             & \texttt{abandon} & The task is very unlikely to be completed within the remaining budget. \\
  Q-finish   & \texttt{finish} & Evidence in the transcript shows the explicit requirements are met. \\
             & \texttt{verify} & Requirements are unchecked or evidence is missing. \\
  \bottomrule
\end{tabularx}
\end{center}
The question instructions were ``The state describes an autonomous terminal agent working on a task at a decision point. Choose the option that best describes the agent's situation.'' (Q-progress) and ``The state describes an autonomous terminal agent that has asked to finish its task. Choose whether the transcript shows the task is complete.'' (Q-finish).
The small LLM received these, the option list and the state JSON, and was told to respond only with \texttt{\{"choice": "<option>"\}}.

\section{Parameters}
\label{app:params}

\begin{center}\small
\begin{tabularx}{\linewidth}{@{}lX@{}}
  \toprule
  Component & Setting \\
  \midrule
  Reasoning model & \texttt{anthropic/claude-sonnet-5.5}; reasoning effort high; 32{,}000 max output tokens; prompt caching \\
  Small LLM & \texttt{anthropic/claude-haiku-5.5}; reasoning disabled; JSON-schema output \\
  Decision model & \texttt{typesafe/jev-1.13} via OpenRouter's System One endpoint \\
  Tools & \texttt{bash(command, timeout\_s)} (default 120\,s, max 1{,}800\,s), \texttt{finish(summary)}; output truncated to 4{,}000 + 8{,}000 characters \\
  Limits (per trial) & 60 steps; \$1.00; 30M tokens; 4\,h wall; 30 controller calls; Harbor agent timeout per task \\
  Triggers & T1: 2 consecutive failures (non-zero exit or timeout); T2: same normalized command 2$\times$ in last 5, or same non-empty output 2$\times$ in last 3; T4: every 10 steps; one trigger per step (priority T1, T2, T4); cooldown 3 steps \\
  Guards & G2: step $\geq 0.4\times$ max and ($\geq$3 consecutive failures or $\geq$2 T2 firings); G3: 2 verifications; G4: timeout 2\,s (Jev), 10\,s (small LLM) \\
  Compact state & instruction $\leq$3{,}000 chars; last 5 actions (command $\leq$300, output tail $\leq$500); finish summary $\leq$1{,}000; total $\leq$24{,}000 chars \\
  Infrastructure & Harbor 0.24.0; Daytona sandboxes from each task image with the task's resources; 20 trials in parallel \\
  \bottomrule
\end{tabularx}
\end{center}

\section{Per-task results}
\label{app:pertask}

Within strata, the C1 pass-rate difference was $+7.1$\,pp (easy), $-22.2$\,pp (mid) and $-8.3$\,pp (hard), and the cost ratio 1.29, 1.48 and 0.76; the mid and hard strata are too small for inference.

\begin{center}\small
\setlength{\tabcolsep}{4pt}
\begin{tabular}{llrrrrrrr}
\toprule
& & \multicolumn{5}{c}{Passes out of 3} & USD/attempt & Cost ratio \\
\cmidrule(lr){3-7}
Task & Stratum & native & rules & small & Jev & placebo & native & Jev/native \\
\midrule
\texttt{build-pmars} & easy & 3 & 3 & 3 & 3 & 3 & 0.025 & 1.14 \\
\texttt{code-from-image} & easy & 3 & 3 & 3 & 3 & 3 & 0.024 & 1.99 \\
\texttt{fix-ocaml-gc} & easy & 3 & 3 & 3 & 3 & 3 & 0.165 & 0.98 \\
\texttt{git-leak-recovery} & easy & 3 & 3 & 3 & 3 & 3 & 0.011 & 0.99 \\
\texttt{headless-terminal} & easy & 3 & 3 & 3 & 3 & 3 & 0.029 & 1.34 \\
\texttt{install-windows-3.11} & easy & 0 & 1 & 3 & 3 & 1 & 0.502 & 1.59 \\
\texttt{modernize-scientific-stack} & easy & 3 & 3 & 3 & 3 & 3 & 0.010 & 1.18 \\
\texttt{path-tracing-reverse} & easy & 3 & 3 & 3 & 3 & 3 & 0.349 & 1.42 \\
\texttt{protein-assembly} & easy & 3 & 2 & 2 & 3 & 3 & 0.150 & 1.48 \\
\texttt{regex-log} & easy & 3 & 3 & 3 & 3 & 3 & 0.032 & 1.91 \\
\texttt{reshard-c4-data} & easy & 3 & 3 & 3 & 3 & 3 & 0.062 & 0.98 \\
\texttt{sqlite-db-truncate} & easy & 3 & 3 & 3 & 3 & 3 & 0.037 & 1.05 \\
\texttt{winning-avg-corewars} & easy & 3 & 3 & 3 & 3 & 3 & 0.287 & 1.28 \\
\texttt{write-compressor} & easy & 3 & 3 & 3 & 3 & 3 & 0.096 & 1.25 \\
\texttt{filter-js-from-html} & mid & 3 & 2 & 0 & 1 & 1 & 0.039 & 1.57 \\
\texttt{gcode-to-text} & mid & 1 & 2 & 2 & 1 & 1 & 0.421 & 1.38 \\
\texttt{git-multibranch} & mid & 2 & 3 & 3 & 2 & 3 & 0.041 & 1.49 \\
\texttt{caffe-cifar-10} & hard & 1 & 0 & 1 & 0 & 0 & 0.141 & 0.78 \\
\texttt{dna-insert} & hard & 0 & 0 & 1 & 0 & 0 & 0.086 & 1.11 \\
\texttt{extract-moves-from-video} & hard & 0 & 0 & 0 & 0 & 0 & 0.265 & 0.43 \\
\texttt{make-doom-for-mips} & hard & 0 & 0 & 0 & 0 & 0 & 0.947 & 0.90 \\
\bottomrule
\end{tabular}

\end{center}

\section{Deviations from the pre-specified plan}
\label{app:deviations}

\begin{center}\small
\begin{tabularx}{\linewidth}{@{}lX@{}}
  \toprule
  Planned & Done, and why \\
  \midrule
  30--40 held-out tasks & 21; calibration covered 30 of 86 eligible tasks \\
  Caps at p90 of passing calibration runs & 60 steps and \$1.00, above every calibration run; the p90 caps (20 steps, \$0.25) would have cut 5 of 44 passing runs and acted as an abandon policy \\
  Provider pinning & Hosts chosen by OpenRouter; serving host logged per call \\
  Temperature 0 & Not sent; Sonnet~5.5 endpoints do not accept it \\
  Original verification feature & Widened before development to cover running script files and test targets, and edits and checks in one command; the original was false at every calibration finish request \\
  Open-weight replication (C5) & Not run: no single quantization across hosts; the Holm family is C2--C4 \\
  \bottomrule
\end{tabularx}
\end{center}

\end{document}